\documentclass[openacc]{rsproca_new}

\usepackage{lipsum}
\usepackage[colaction]{multicol}
\usepackage{lineno}
\usepackage{setspace} 
\usepackage{float}
\usepackage{placeins}

\DeclareUnicodeCharacter{0301}{\'e}

\jname{rsif}
\Journal{J R Soc Interface\ }

\begin{document}
\title{Predicting the long-term collective behaviour of fish pairs with deep learning}

\author{
Vaios Papaspyros$^{1}$, Ram\'on Escobedo$^{2}$, Alexandre Alahi$^{3}$, Guy Theraulaz$^{2}$, Cl\'ement Sire$^{4}$, Francesco Mondada$^{1}$
}

\address{$^{1}$Mobile Robotic Systems (Mobots) group, Institute of Electrical and Micro Engineering, \'Ecole Polytechnique F\'ed\'erale de Lausanne (EPFL), CH-1015 Lausanne, Switzerland\\
$^{2}$Centre de Recherches sur la Cognition Animale, Centre de Biologie Int\'egrative, CNRS, Universit\'e de Toulouse III -- Paul Sabatier, 31062 Toulouse, France\\
$^{3}$VITA group, Civil Engineering Institute, \'Ecole Polytechnique F\'ed\'erale de Lausanne (EPFL), CH-1015 Lausanne, Switzerland\\
$^{4}$Laboratoire de Physique Th\'eorique, CNRS, Universit\'e de Toulouse III -- Paul Sabatier, 31062 Toulouse, France 
}

\subject{behavioural biology, machine learning, biophysics}


\corres{Vaios~Papaspyros (\email{Vaios.Papaspyros@epfl.ch})\\
Cl\'ement~Sire (\email{Clement.Sire@univ-tlse3.fr})
\\
}




\begin{abstract}
    \begin{linenumbers}
        Modern computing has enhanced our understanding of how social interactions shape collective behaviour in animal societies. Although analytical models dominate in studying collective behaviour, this study introduces a deep learning model to assess social interactions in the fish species \textit{Hemigrammus rhodostomus}. We compare the results of our deep learning approach to experiments and to the results of a state-of-the-art analytical model. To that end, we propose a systematic methodology to assess the faithfulness of a collective motion model, exploiting a set of stringent individual and collective spatio-temporal observables. We demonstrate that machine learning models of social interactions can directly compete with their analytical counterparts in reproducing subtle experimental observables. Moreover, this work emphasises the need for consistent validation across different timescales, and identifies key design aspects that enable our deep learning approach to capture both short- and long-term dynamics. We also show that our approach can be extended to larger groups without any retraining, and to other fish species, while retaining the same architecture of the deep learning network. Finally, we discuss the added value of machine learning in the context of the study of collective motion in animal groups and its potential as a complementary approach to analytical models.
    \end{linenumbers}
\end{abstract}


\begin{fmtext}

\section{Introduction}

Collective behaviour in animal groups is a very active field of research, studying the fundamental mechanisms by which individuals coordinate their actions~\cite{sumpter2010collective, krause2002living, ball2011flow} and self-organise~\cite{camazine2020self, couzin2003self}. One of the most common forms of collective behaviour can be observed in schools of fish and flocks of birds that have the ability to coordinate their movements to collectively escape predator attacks or improve their foraging efficiency~\cite{vicsek2012collective, cavagna2018physics}. This coordination at the group level mainly results from the social interactions between individuals. Important steps to understand these collective phenomena consist in characterising these interactions and understanding the way individuals integrate interactions with other group members~\cite{deutsch2020multi, herbert2011inferring, herbert2016understanding, gautrais2012deciphering, calovi2018disentangling}.

\end{fmtext}


\maketitle

\begin{multicols}{2}

New tracking techniques and tools for behavioural analysis have been developed that have greatly improved the quality of collective motion data~\cite{branson2009high, dell2014automated, gallois2021fasttrack, anderson2014toward, perez2014idtracker, romero2019idtracker, walter2021trex}. In particular, advances in computing have allowed the development of computationally demanding data-oriented model generation techniques~\cite{calovi2018disentangling, escobedo2020data, jayles2020collective, cazenille2019automatic, heras2018deep, costa2020automated} and the subsequent simulation of biological models~\cite{gilpin2020learning}. This has resulted in more realistic models that attempt to recover the social interactions that govern collective behaviours. Yet, the bottleneck with most of these approaches is that they rely on demanding and laborious mathematical work to obtain the interactions from experimental data.

An alternative to such analytical models is to exploit machine learning (ML) techniques and let an algorithm learn the interactions directly from data. The know-how required to use these techniques is different from the one needed to design analytical models. Nevertheless, the structure of ML algorithms, here a neural network, has an impact on the modelling performance, and requires specific expertise~\cite{mammadli2019art}. Once the architecture of an ML algorithm is set, ML can often process data for different species without structural adaptation, and generate new models quickly. This is very different from analytical models, where each new species requires redefining the model nearly from scratch. The downside of this flexibility is that ML models are usually less explainable (``black box''). Yet, recent ML algorithms provide higher-level information mappable to more tangible formats, such as force maps, which show the strength and direction of behavioural changes experienced by an individual when interacting with other individuals in a moving group~\cite{heras2018deep, costa2020automated}. Despite their limited explainability, ML algorithms require only a few biological assumptions. They offer an almost hypothesis-free procedure~\cite{valletta2017applications} that can even outperform human experts in detecting subtle patterns~\cite{marques2018structure}, making ML a very appealing complementary approach to analytical models.

For both analytical and ML models, several studies evaluate models over \textit{short timescales} and through instantaneous quantities such as speed, acceleration, distance and angle to objects~\cite{cazenille2018blend, cazenille2019automatic}, or by measuring the error between predictions and ground truth~\cite{heras2018deep, alahi2016social, kothari2020human}. Only more recently, long timescales have also been considered~\cite{jayles2020collective}. However, a model that performs well at short timescales compared to experiment does not necessarily perform well at long timescales. This is especially true for models that try to reproduce complex collective phenomena in living systems. To our knowledge, the predictive capacity of ML models in this context has not been evaluated over both short \textit{and} long timescales, that is, their ability to generate synthetic data that replicate the outcomes of social interactions over both timescales.

Here, we demonstrate that ML models can generate realistic synthetic data with minimal biological assumptions, and that they allow to accelerate and generalise the process of collective behaviour modelling. More specifically, we present a social interaction model using a deep neural network that captures both the short- and long-term dynamics observed in schooling fish. We apply our approach to pairs of rummy-nose tetra (\textit{Hemigrammus rhodostomus}) swimming in a circular tank, and show that it can also be applied to fish species with similar burst-and-coast swimming (zebrafish; \textit{Danio rerio}). Our ML model is benchmarked against the state-of-the-art analytical model for this species~\cite{calovi2014swarming}, showing that it performs as well as the latter, even for very subtle quantities measured in the experiments. Moreover, we also introduce a systematic methodology to stringently test the results of an analytical or ML model against experiment, at different timescales, and in the context of animal collective motion.
    
\section{Methods}

\label{sec:materials}

\subsection{Experimental data}

The trajectory data used in this study were originally published in~\cite{calovi2018disentangling} for \textit{Hemigrammus rhodostomus} swimming either alone or in pairs in a circular tank of radius 25\,cm. This species is characterised by a burst-and-coast swimming mode, where the fish perform a succession of sudden and short acceleration periods (of typical duration 0.1\,s), each followed by a longer gliding period almost in a straight line, resulting in a mean total duration of the kicks of 0.6\,s. The instant of the kicks, when heading changes take place, are assimilated to decision instants~\cite{calovi2018disentangling}.

The dataset corresponds to 15 hours of video recordings at 25\,Hz. Fish are tracked with idTracker~\cite{perez2014idtracker}, an image analysis software which extracts the 2D trajectories of all individuals. Occasionally, the tracking algorithm is temporarily unable to report positions accurately. This can be due to small fluctuations in lighting conditions, fish standing still or moving at very low speed, fish swimming very close to the surface, to the border, or to each other. These instances are corrected using several filtering processes. Since our analyses focus on social interactions, we remove the periods during which fish are inactive. Fish body length (BL) is of about 3.5\,cm, and the intervals of time during which fish velocity is less than 1\,BL/s are removed. Large leaps in fish trajectories during which fish move by more than 1.5\,BL~$\approx 5.25$\,cm between two consecutive frames, meaning that fish move at almost 65\,cm/s, are also identified and removed, as they result from tracking errors. Finally, missing points are filled by linear interpolation. The final dataset used in this work represents approximately 4 hours of trajectory data for pairs of \textit{H.~rhodostomus}. 

Moreover, trajectories of the original dataset have been resampled with a timestep of $\Delta t = 0.12$\,s instead of the original 0.04\,s provided by the camera, and data points have been converted from pixel space to a normalised $[-1, 1]$ range to facilitate the training of the networks. This subsampling helps to reduce the random noise between subsequent camera frames at the very short timescale of 0.04\,s  (especially for measuring fish headings and speeds), while maintaining a sufficiently small timestep to study and model the social interactions. The new timestep $\Delta t = 0.12$\,s is of the same order as the sudden acceleration period of a kick and approximately one fifth of the average total kick duration \cite{calovi2018disentangling}. In addition to reducing the noise, the subsampling also reduces the dimension of the input vector and of the effective size of the training dataset and, as a result, of the training time for the ANN models presented in this work.

\begin{figure*}[t!]
    \centering
    \includegraphics[width=1.0\linewidth]{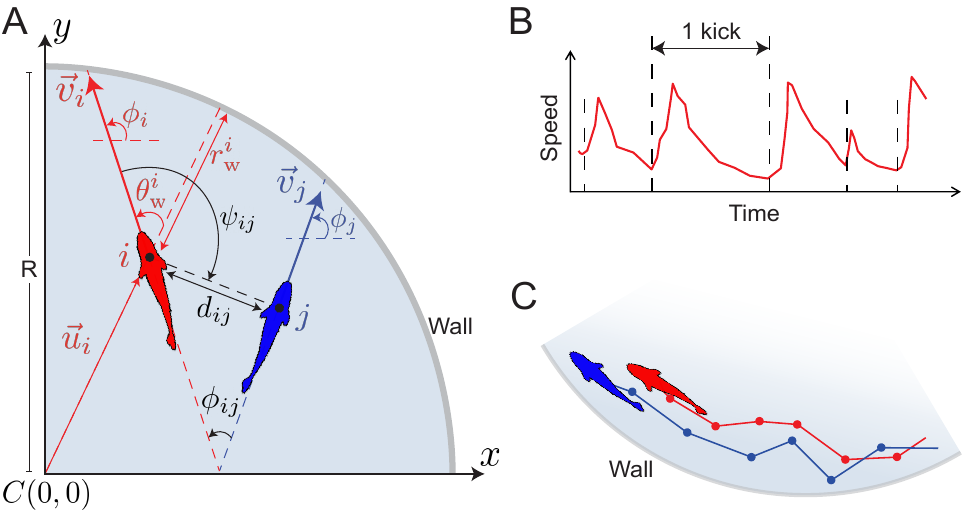}
    \caption{{\bf A.} Individual and collective variables characterising the instantaneous state of an individual (focal fish in red) and its pairwise relation with a neighbour (blue): distance to the wall $r_{\rm w}^i(t)$, angle of incidence to the wall $\theta_{\rm w}^i(t)$, heading angle $\phi_i(t)$, distance between individuals $d_{ij}(t)$, difference of heading angles $\phi_{ij}(t)$, and angle of perception $\psi_{ij}(t)$. Positive angles (curved arrows) are defined in the anti-clockwise direction, starting from the positive semi-axis of abscissas. The radius of the circular setup is $R=25$\,cm. For visualisation purposes, the size of fish is not to scale with the tank. {\bf B.} Typical profile of the fish speed, $V(t)$, showing the typical sequence of kicks (abrupt accelerations followed by longer gliding phases). {\bf C.} Trajectories of two fish close to the wall due to their burst-and-coast swimming mode. The dots in the trajectories denote the instants of the kicks, where fish decision-making is assumed to take place.}
    \label{fig:quantities_viz}
\end{figure*}

\subsection{Quantification of individual and collective behaviour in pairs of fish}

We use a set of observables to quantify how close the results of the models are from the measures obtained in the experiments~\cite{calovi2018disentangling, escobedo2020data, jayles2020collective}. These observables constitute a stringent benchmarking and validation  when designing and testing a model. In the case of deep learning techniques, those observables also serve as means to partially explain what the algorithm has learned.

Let us first define the temporal variables characterising the individual and collective behaviour of the fish.
Fig.~\ref{fig:quantities_viz}A shows two fish swimming in a circular tank of radius $R=25$\,cm. The position vector of a fish $i$ at time $t$ is given by its Cartesian coordinates $\vec{u}_i(t)=(x^i(t),y^i(t))$ in the system of reference, centred at the centre of the tank $C(0,0)$. The components of the velocity vector $\vec{v}_i(t)=(v_x^i(t),v_y^i(t))$ are given by $v_x^i(t)=(u_x^i(t)-u_x^i(t-\Delta t))/\Delta t$ (similar expression for $v_y^i$). The heading angle of the fish is assumed to indicate its direction of motion and is therefore given by the angle that the velocity vector forms with the horizontal, $\phi_i(t) = {\tt ATAN2}(v_y^i(t),v_x^i(t))$.

The motion of a given fish~$i$ is then described using the three following instantaneous variables: the speed, $V_i(t) = \| \vec{v}_i(t) \|$, the distance of the fish to the wall, $r_{\rm w}^i(t) = R - \| \vec{u}_i(t) \|$, and the angle of incidence of the fish to the wall, $\theta_{\rm w}^i(t)$, defined by the angle formed by the velocity vector and the normal to the wall: $\theta_{\rm w}^i(t)=\phi_i(t) -{\tt ATAN2}(y^i(t),x^i(t))$, see Fig.~\ref{fig:quantities_viz}A.

When there are two fish $i$~and~$j$ in the tank, their relative motion is characte\-rised by means of three variables: the distance between fish, $d_{ij}(t) = \| \vec{u}_i(t) - \vec{u}_j(t) \|$, the difference between their heading angles, $\phi_{ij}(t) = \phi_j(t) - \phi_i(t)$, which measures the degree of alignment between both fish, and the angle of view, $\psi_{ij}(t)$, which is the angle with which fish~$i$ perceives fish~$j$, and which is generally independent of $\psi_{ji}(t)$. See Fig.~\ref{fig:quantities_viz}A for the graphical representation of these quantities. The angle of perception of the fish also allows us to define the notion of \textit{geometrical leadership} for two fish: fish~$i$ is the \textit{geometrical leader} (and therefore, $j$~is the \textit{geometrical follower}), if $\lvert \psi_{ij}(t) \rvert > \lvert \psi_{ji}(t) \rvert$, meaning that $i$~has to turn by a larger angle to face~$j$ than the angle that $j$~has to turn to face~$i$. In practice, these definitions of the geometrical leader and follower provide a precise and intuitive characterisation of a fish being ahead of the other. Note that being the leader or the follower is an instantaneous state that can change from one kick to the other.

These 6 quantities $V_i(t)$, $r_{\rm w}^i(t)$, $\theta_{\rm w}^i(t)$, $d_{ij}(t)$, $\phi_{ij}(t)$, and $\psi_{ij}(t)$ being defined, the measure of their probability distribution functions (PDF) constitutes a set of observables probing the individual and collective instantaneous fish dynamics in a fine-grained and precise manner. The PDF of $V_i(t)$, $r_{\rm w}^i(t)$, $\theta_{\rm w}^i(t)$ probe the behaviour of a focal fish sampled over the observed dynamics, and are hence called \textit{instantaneous individual observables}. The PDF of $d_{ij}(t)$, $\phi_{ij}(t)$, and $\psi_{ij}(t)$ characterise the correlations between 2 fish \textit{at the same time} $t$ and are hence called \textit{instantaneous collective observables}. These 3 collective observables can be easily generalised to a group of arbitrary size $N>2$, by considering $i$ and $j$ as pairs of nearest neighbours, or pairs of second-nearest neighbours (or even farther neighbours), or even averaging them over all pairs in the group (then probing the size, the polarisation, and the anisotropy of the group). Ultimately, comparing experimental results and model predictions for these individual and collective observables constitutes a stringent test of a model.

Moreover, to characterise the \textit{temporal correlations} arising in the dynamics, we make use of 3~additional observables involving quantities measured \textit{at two different times}, for a given focal fish~\cite{jayles2020collective}: the mean-squared displacement $C_X(t)$, the velocity autocorrelation $C_V(t)$, and, especially challenging, the autocorrelation of the angle of incidence to the wall $C_{\theta_{\rm w}}(t)$, defined respectively by
\begin{align}
    C_X(t)                & = \left\langle\left[\vec{u}_i(t + t') - \vec{u}_i(t')\right]^ 2\right\rangle,
    \label{eq:cx}
    \\
    C_V(t)                & = \left\langle \vec{v_i}(t + t') \cdot \vec{v}_i(t')\right \rangle,
    \label{eq:cv}
    \\
    C_{\theta_{\rm w}}(t) & = \left\langle\cos\left[\theta_{\rm w}^i(t + t') - \theta_{\rm w}^i(t')\right]\right \rangle,
    \label{eq:ctheta}
\end{align}
where $\langle w(t) \rangle$ is the average of a variable $w(t)$ over all reference times $t'$ (assumption of a stationary dynamics, where correlations between two times depend solely on their time separation), over all focal fish, and over all experimental runs. 
In principle, these correlation observables can also be generalised to probe the (collective) time correlations between the two different fish (or between nearest neighbours in a group of $N>2$ individuals). For instance, one could consider $C_{V_{NN}}^{\rm }(t) = \left\langle \vec{v_i}(t + t') \cdot \vec{v}_j(t')\right \rangle$, where the average is now over nearest neighbour pairs. However, in the present study, we will limit ourselves to the study of the 3 (individual) correlation functions listed in Eqs.~(\ref{eq:cx}-\ref{eq:ctheta}).

\subsection{Analytical and deep learning models of fish behaviour}
\label{sec:abc}

Many species of fish like \textit{H.~rhodostomus} or \textit{Danio rerio} move in a \textit{burst-and-coast} manner, meaning that their swimming pattern consists of a sequence of abrupt accelerations each followed by a longer gliding period (Fig.~\ref{fig:quantities_viz}B), during which a fish moves more or less in a straight line (Fig.~\ref{fig:quantities_viz}C). The kicking instants observed in the curve of the speed can be interpreted as decision times when the fish potentially initiates a change of direction. In \textit{H.~rhodostomus}, the mean time interval between kicks and the typical kick length were experimentally found to be close to 0.5\,s and~7\,cm, respectively~\cite{calovi2018disentangling}. When confined in circular tanks, fish tend to swim close to the curved wall because their trajectory is made of quasi straight segments with limited variance of the heading angle between kicks, hence preventing the fish to escape from the tank wall (unless when a rare large heading angle change occurs)~\cite{calovi2018disentangling, xue2023tuning}. When swimming in groups, \textit{H.~rhodostomus} tend to remain close to each other, especially when the number of fish in the tank is small. In fact, the social interactions between fish reflect the combined tendency to align with and follow their neighbours while at the same time maintaining a safe distance with the wall. At a given kicking instant, only a few neighbours (one or two) have a relevant influence on the behaviour of a fish~\cite{lei2020computational}. The decision-making of fish displaying a burst-and-coast swimming mode can thus be reproduced by considering only pairwise interactions. Obviously, if one only considers pairs of fish, like here, it therefore suffices to consider the relative state of the neighbouring fish (relative position and velocity) and the effect of the distance and the relative orientation to the wall~\cite{calovi2018disentangling, escobedo2020data}.

\begin{figure*}[!ht]
    \centering
    \includegraphics[width=1.\linewidth]{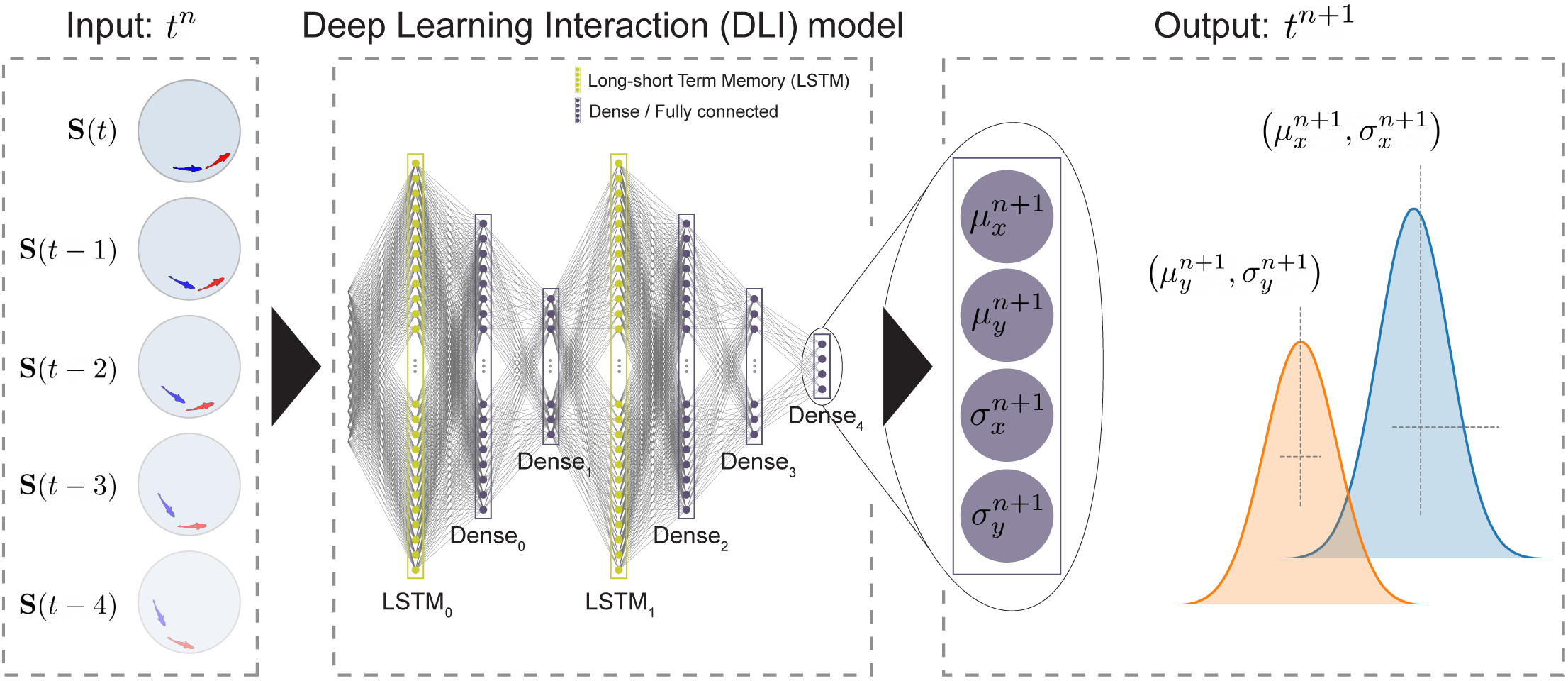}
    \caption{Structure of the Artificial Neural Network (ANN) used in the Deep Learning Interaction (DLI) model.
        From left to right: \textit{Input} of the ANN: the 5 last states, $({\bf S}(t-4), \dots, {\bf S}(t))$ at time $t$. Where ${\bf S}(t) = \big( {\bf s}_i(t), {\bf s}_j(t), d_{ij}(t) \big)
            \in \mathbb{R}^{11}$ and each state is parametrised as ${\bf s}_i(t) = \big(\vec{u}_i(t), \vec{v}_i(t), r_{\rm w}^i(t) \big) \in \mathbb{R}^5$; the 7 layers (two Long-short Term Memory, also known as LSTM, layers and 5 Dense Layers) capturing the social dynamics; \textit{Output}: the two pairs of values $(\mu_x,\sigma_x)$ and $(\mu_y,\sigma_y)$ corresponding respectively to the mean and standard deviation of the probability distribution function (assumed to be Gaussian) of each component $a_x$ and $a_y$ of the instantaneous acceleration vector $\vec{a}$ at time $t+1$, constituting the prediction of the DLI model.}
    \label{fig:nn_structure}
\end{figure*}

\subsubsection{Analytical Burst-and-Coast model}
\label{sec:burst-and-coast}

The Analytical Burst-and-Coast model (hereafter called ABC model) quantitatively reproduces the dy\-na\-mics of \textit{H.~rhodostomus} swimming alone or in pairs under the hypothesis that fish decision-making times correspond exactly to their kicking times, that is, the new direction of movement, the duration, and the length of the kick are decided precisely at the end of the previous kick~\cite{calovi2018disentangling}.

Given a pair of agents $i$ and $j$ at a respective state $(\vec{u}_j^{\,n},\phi_j^n)$ and $(\vec{u}_i^{\,n},\phi_i^n)$ at time $t^n$, the state of agent~$i$ at the next instant of time $t_i^{n+1}$ is given by
\begin{eqnarray}
    && t_i^{n+1} = t_i^n + \tau_i^n, \label{ABC-t}\\
    && \phi_i^{n+1} = \phi_i^n + \delta \phi_i^n, \label{ABC-phi}\\
    && \vec{u}_i^{\,n+1} = \vec{u}_i^{\,n} + l_i^n \, \vec{e}\,(\phi_i^{n+1}), \label{ABC-u}
\end{eqnarray}
where $\vec{e}\,(\phi_i^{n+1})$ is the unitary vector pointing in the heading direction~$\phi_i^{n+1}$, $\tau_i^n$ and $l_i^n$ are the duration and length of the $n$-th kick of agent~$i$, and $\delta \phi_i^n$ is the heading change of agent~$i$. The heading angle change $\delta \phi_i^n$ is the result of three effects: the interaction with the wall, the social interactions with the other fish (repulsion/attraction and alignment), and the natural spontaneous fluctuations of fish headings (cognitive noise)~\cite{calovi2018disentangling}. 
The term “cognitive noise” encapsulates the fact that fish (or humans) would not generally replicate the exact same motion when placed under identical initial conditions, namely starting at the same positions and with the same initial velocities. Hence, a behavioural model must not only describe the social interactions between individuals, but also the properties of their spontaneous fluctuations.
The social interactions depend only on the relative state of both agents, determined by the triplet $(d_{ij},\psi_{ij},\phi_{ij})$. The derivation of the shape and intensity of the functions involved in $\delta \phi_i^n$ is based on phy\-si\-cal principles of symmetry of angular functions and a data-driven reconstruction procedure detailed in~\cite{calovi2018disentangling} for the case of \textit{H.~rhodostomus} and in~\cite{escobedo2020data} for the general case of animal groups.

Starting from the initial condition $(\vec{u}_i^{\,0},\phi_i^0)$ of fish~$i$, the length and the duration of its next kick, $l_i^0$ and $\tau_i^0$, are sampled from the experimental distributions obtained in~\cite{calovi2018disentangling}. Then, the timeline $t_i^1$ of fish~$i$ is updated with Eq.~(\ref{ABC-t}), the heading angle of the next kick $\phi_i^1$ is calculated with Eq.~(\ref{ABC-phi}), and the position of the fish at the end of the kick $\vec{u}_i^1$ is obtained with Eq.~(\ref{ABC-u}). As kicks of different fish are asynchronous, the next kick can be performed by any of the two fish. Each fish has thus it own timeline, but is subject, at each of its kicks, to the evolution of the other fish along its own kicks.

The ABC model is therefore a discrete model that generates kick events instead of continuous time positions. To directly compare with the DLI model presented in the next section, which is a continuous time model, we re-sampled the trajectories made of kick events produced by the ABC model and build continuous time trajectories with a timestep of size $\Delta t = 0.12$\,s. We produced trajectories that add up to a total of 500,000 timesteps, corresponding to approximately 16.7 hours.


\subsubsection{Deep Learning Interaction model}
\label{sec:ann_structure}

The Deep Learning Interaction model (hereafter called DLI model) consists of an Artificial Neural Network (ANN) which is fed with a set of variables characterising the motion of \textit{H.~rhodostomus} and which provides the necessary information to reproduce the social interactions of these fish by estimating their motion along timestep of length $\Delta t = 0.12\,s$. At time~$t$, the DLI model is designed to take sequences of states as input to capture the short- and long-term dynamics. Then, it generates predictions for the acceleration components of the fish at the following timestep $t+\Delta t$.

For the DLI model, the state of an agent~$i$ at time $t$ is defined by
\begin{equation}
    {\bf s}_i(t) = \big(\vec{u}_i(t), \vec{v}_i(t), r_{\rm w}^i(t) \big)
    \in \mathbb{R}^5.
    \label{eq:ind_state}
\end{equation}
The state of an agent includes redundant information: in a fixed geometry, $r_{\rm w}^i$ can be deduced from $\vec{u}_i$, and $\vec{v}_i^{\,n}$ from the input sequence $\vec{u}_i^{\,n-4}, \dots, \vec{u}_i^{\,n}$. This redundancy is intended to facilitate the training process of the neural network. Furthermore, these redundancies are shown to significantly boost the performance of the network compared to similar ANN structures (see \nameref{SIText1}).

The system's state ${\bf S}(t)$ is then defined as the combination of both agent states, in addition to their inter-individual distance $d_{ij}(t)$ (also a redundant variable):
\begin{equation}
    {\bf S}(t) = \big( {\bf s}_i(t), {\bf s}_j(t), d_{ij}(t) \big)
    \in \mathbb{R}^{11}.
    \label{eq:input_vec}
\end{equation}

Fig.~\ref{fig:nn_structure} shows the structure of the ANN, consisting of 7~layers: two Long-Short Term Memory (LSTM) layers~\cite{hochreiter1997long}, and 5 fully connected (Dense) layers.

The first LSTM layer consists of 256 neurons and is located at the input of the ANN, where it receives the sequence of the 5 last states of the system, \textit{i.e.}, a matrix of dimension~$5\times 11$: $({\bf S}(t-4), \dots, {\bf S}(t))$. This history length of 4~timesteps (0.48\,s) is borrowed from the biology of the fish: as already mentioned, the time it takes for a fish to display its characteristic behaviour, a kick, is 0.5\,s~\cite{calovi2018disentangling}, therefore, we input the current state plus the states that correspond to the average duration of a kick. The output of the first LSTM is then gradually reduced in dimension by two successive dense layers, and then scaled up again with a second LSTM, whose configuration is also based on a history of 5 states. Then, two other dense layers are used to reduce the dimension of the output of the second LSTM, and a last dense layer is applied to provide the final output of the ANN. More details about the configuration of the ANN are given in Table~S7 in \nameref{SIText1}.

The output of the ANN consists of two pairs of values, $(\mu_x,\sigma_x)$ and $(\mu_y,\sigma_y)$, corresponding to the expected value and standard deviation of the $x$ and $y$ components of the predicted acceleration, which are assumed to be Gaussian distributed~\cite{chua2018deep}, as actually found for \textit{H.~rhodostomus}~\cite{calovi2018disentangling}. Hence, the predicted acceleration of the agent, $\vec{a} = (a_x,a_y)$, can be written
\begin{equation}
    a_x = \mu_x+\sigma_x g_x, \quad a_y = \mu_y+\sigma_x g_y,\label{eq:stoch_accel}
\end{equation}
where $g_x$ and $g_y$ are independent standard Gaussian random variables drawn from ${\cal N}(0,1)$.
Then, the velocity vector of the agent~$i$ at the time $t^{n+1}$ is given by
\begin{equation}
    \vec{v}_i^{\,n+1} = \vec{v}_i^{\,n} + \Delta t \, \vec{a}_i^{\,n},
    \label{eq:motion_eqs1}
\end{equation}
and the position of the agent is updated according to
\begin{equation}
    \vec{u}_i^{\,n+1} = \vec{u}_i^{\,n} + \Delta t \, \vec{v}_i^{\,n+1}.
    \label{eq:motion_eqs2}
\end{equation}

Note that in the DLI model, the predicted variance of the acceleration accounts for the fish intrinsic spontaneous behaviour exhibited during their decision process (cognitive noise), and hence translates the fact that 2 real (or modelled) fish will not act the same if put twice in the same given state characterised by Eq.~(\ref{eq:input_vec}).

In some rare instances, the prediction of the DLI model would move one or both fish outside the limits of the tank. To account for that, we introduce a rejection procedure: the invalid move is rejected, and we resample the Gaussian random variables drawn in Eq.~(\ref{eq:stoch_accel}) until a valid move is produced. Note that a similar rejection procedure is also implemented in the ABC model of \cite{calovi2018disentangling}, to strictly enforce the presence of the wall. Indeed, in the ABC model, the ABC agents would systematically escape the tank after a few seconds or very few minutes without this rejection procedure. In section~\ref{sec:results}\ref{sec:complementary} and Fig.~S1 and S2 in \nameref{SIText1}, we show that the DLI model has, in fact, implicitly learned the presence of the wall, and that DLI agents can remain within or in the close vicinity of the tank for several dozen of minutes without implementing this rejection procedure (60\,\% chance not to escape the tank during 100 minutes of simulation).

The \textit{prediction} of the ANN for at time $t^{n+1}$ is thus a vector of dimension~$1\times 4$ that can be written as
$(\vec{\mu}^{\,n+1}_{\rm pred},\vec{\sigma}^{\,n+1}_{\rm pred})$, where
\begin{equation}
    \vec{\mu}_{\rm pred}^{\,n+1} = \big(\mu_x^{n+1},\mu_y^{n+1}\big)
    \quad \mbox{and} \quad
    \vec{\sigma}_{\rm pred}^{\,n+1} = \big(\sigma_x^{n+1}, \sigma_y^{n+1}\big).
    \label{eq:output_vec}
\end{equation}

The ANN is then trained to approach the \textit{real/observed} values $\vec{\mu}_{\rm real}^{\,n+1}$ by means of the Adaptive Moment Estimation Optimiser (Adam) with a time-decaying learning rate \mbox{$\lambda = 10^{-4}$} and a negative log-likelihood loss function $\ell$ defined in terms of the prediction error $\vec{\epsilon}_{n+1} = \vec{\mu}_{\rm pred}^{\,n+1} - \vec{\mu}_{\rm real}^{\,n+1}$ and the standard deviations as follows~\cite{kingma2014adam}:
\begin{multline}
    \ell(\vec{\epsilon}_{n+1},\vec{\sigma}^{n+1})  = \\ \frac{1}{2} \sum_{n=1}^{N_h}
    \left[\vec{\epsilon}_{n+1}\right]^T {\cal C}^{-1}(\vec{\sigma}^{n+1})
    \, \vec{\epsilon}_{n+1}
    + \frac{N_h}{2} {\rm log} \| {\cal C}(\vec{\sigma}^{n+1}) \|,
    \label{eq:loss}
\end{multline}
where $N_h$ is the number of timesteps in the history of the input of the ANN (here $N_h = 5$) and ${\cal C}$ is a diagonal covariance matrix with the values of $\vec{\sigma}^{n+1}_{\rm pred}$ in the diagonal and zeroes elsewhere.

The training of the ANN is carried out with a subset of the experimental dataset. More specifically, the training process is given a budget of 45 epochs with a batch size of 512 samples on a dataset that was split 80\%, 15\%, and 5\% for training, validation, and test, respectively. Then, the DLI model is used to produce trajectories of 500,000 timesteps of size $\Delta t = 0.12$\,s, as done with the ABC model. At the beginning of the simulation, each agent is given a copy of the DLI model and both agents are initialised with a random 5-timestep-long trajectory sampled from the fish dataset. At each timestep $t^{n}$, the state vector ${\bf S}(t^n)$ is built and introduced in the network, which provides the estimated instantaneous acceleration distributions at time $t^{n+1}$. Then, the acceleration is evaluated according to Eq.~(\ref{eq:stoch_accel}), and the next positions and velocities of the agents are obtained from the equations of motion, Eqs.~(\ref{eq:motion_eqs1}, \ref{eq:motion_eqs2}).

\begin{figure*}[!ht]
    \centering
    \includegraphics[width=1.\linewidth]{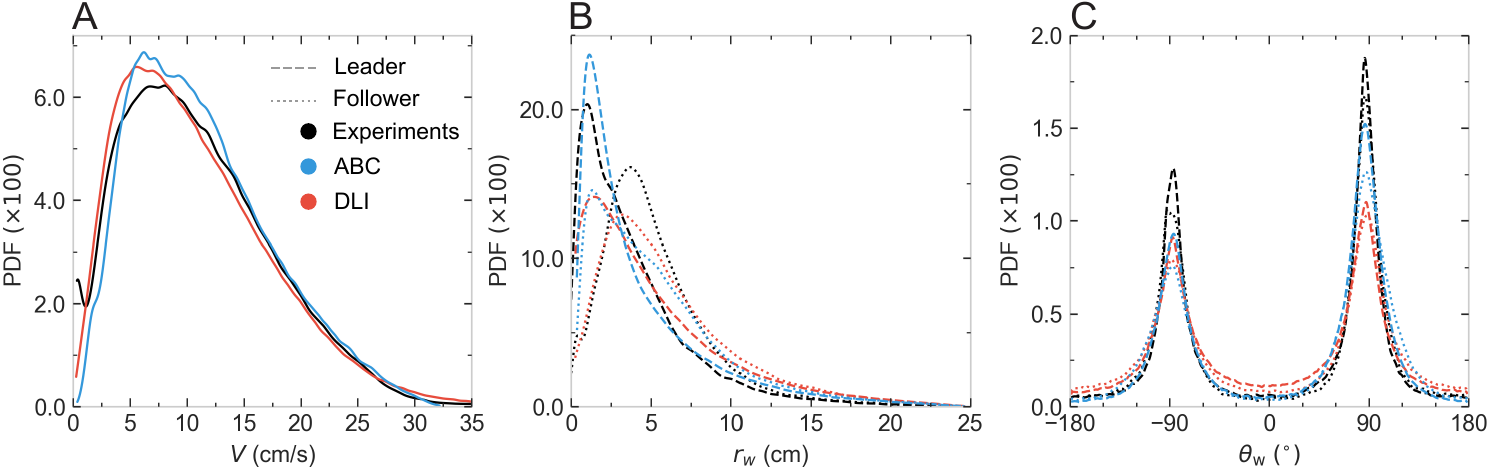}
    \caption{Probability density functions (PDF) of observables characterising individual behaviour: {\bf A} Speed $V$, {\bf B} distance to the wall $r_{\rm w}$, and {\bf C} angle of incidence to the wall $\theta_{\rm w}$. Black lines: experimental fish data. Blue lines: agents of the Analytical Burst-and-Coast model (ABC). Red lines: agents of the Deep Learning Interaction model (DLI). In panels {\bf B} and {\bf C}, dashed lines: geometrical leader; dotted lines: geometrical follower.}
    \label{fig:individual_quantities_virtual}
\end{figure*}

\paragraph{Designing the DLI model}

Designing and selecting an appropriate ANN structure to model a system is for the most part non-trivial and requires either an extensive search through automatic methods (\textit{e.g.}, neuro-evolution~\cite{martin2007evolution, mwaura2015evolving, sekaj2019neuro}) or an exhaustive number of empirical attempts for very specific applications~\cite{cazenille2019automatic, heras2018deep, costa2020automated}. Here, we followed a hybrid approach consisting in empirically designing an ANN based on biological insight and automatically searching for its optimal structure by bootstrapping the search. Once we established this initial model, we performed an automated search for similar neural networks using the same input and output for different combinations of $i)$ the number of layers, $ii)$ the size of the layers, and $iii)$ the activation functions (\textit{i.e.}, transfer functions tasked with mapping the inputs of a neuron to a single weighted output value passed to the next layer). The search included a total of 82 neural network structures, trained with the same budget of iterations and stopping criteria, and out of which the ANN shown above is the best performing. The best performing ANN is selected according to the metrics presented in the following section. 

Three notable categories of networks were considered: $i)$ non-probabilistic networks that only generate $\big(\mu_x^{n+1},\mu_y^{n+1}\big)$ (and hence, not explicitly including the cognitive noise), $ii)$ probabilistic networks that do not have memory cells (hence, missing the fact that fish are gliding passively on a timescale of order 0.5\,s), and $iii)$ probabilistic networks that implement memory thanks to LSTM layers. Non-probabilistic networks $(i)$ provide the mean value of the components of the acceleration for the next timestep with high accuracy, but miss the essential variability that is intrinsic to the spontaneous behaviour of fish and which allows for the emergence of social interactions. Probabilistic networks without memory $(ii)$ are able to partly capture this intrinsic variability, but do not fully capture the non-linear nature of the problem (see Fig.~S6 in \nameref{SIText1} and \nameref{mov:mli}). Finally, probabilistic networks with memory $(iii)$ performed generally well, and we found that the structure used in the DLI model consistently provides the best results for the number of epochs set for training and for the ANNs considered by the automatic search.

Our search approach revealed the existence of two crucial ingredients that must be considered in the model, both accounting for biological characteristics of fish behaviour observed experimentally. First, the neural network must be fed with information covering the typical timescale along which relevant changes take place in the behaviour of the fish. Since real fish kicks last 0.5--0.6\,s on average, the NN needs information about the fish behaviour over time intervals of at least this duration (that is, 4 to 5 timesteps of 0.12\,s). However, we found that using longer vector lengths (up to 10 timesteps) for the case of \textit{H.~rhodostomus} does not lead to any significant improvement in the results, while considerably increasing the training time. Second, the output of the network must contain a sufficiently wide diversity of predictions so that the agents reproduce the high variability of responses that fish display when behaving spontaneously and reacting to external stimuli.

ANNs without memory tend to make too similar predictions, and agents do not initiate the typical direction changes that are observed in the experiments. A possible solution could be to add some phenomenological noise to the predictions of the NN. However, this would result in an unrealistic behaviour, albeit an improvement over not adding noise at all. For example, when a fish swims close to the wall, it does not have the same liberty to turn toward or away from the wall, which would not be captured by a too crude implementation of the fish cognitive noise. Our approach accounts for this behavioural uncertainty for each state (position, velocity, distance to the neighbour and to the wall) and for both degrees of freedom during the training phase of the ANN, being therefore able to capture these complex behavioural patterns. The performance of the two variants is depicted in Fig.~S4 in \nameref{SIText1}.

\section{Results}

\label{sec:results}

When fish swim in a circular tank (here, of radius $R = 25$\,cm),  they interact with each other and with the tank wall. The resulting collective dynamics can be finely characterised by exploiting the 9 observables introduced and described in the Methods section. As explained there, these observables probe 1)~the instantaneous individual behaviour, 2)~the instantaneous collective behaviour, and 3)~the temporal correlations of the dynamics.

Hereafter, we analyse three trajectory datasets: the first one corresponds to pairs of \textit{H.~rhodostomus} in our experiment (4 hours of data), the second one to the Analytical Burst-and-Coast model (ABC; 16.7 hours), and the third one to the Deep Learning Interaction model (DLI; 16.7 hours). \nameref{mov:comparison} shows typical trajectories for these three conditions. The aim of this section is to quantitatively validate the qualitative agreement observed in this video.

\begin{figure*}[!ht]
    \centering
    \includegraphics[width=1.\linewidth]{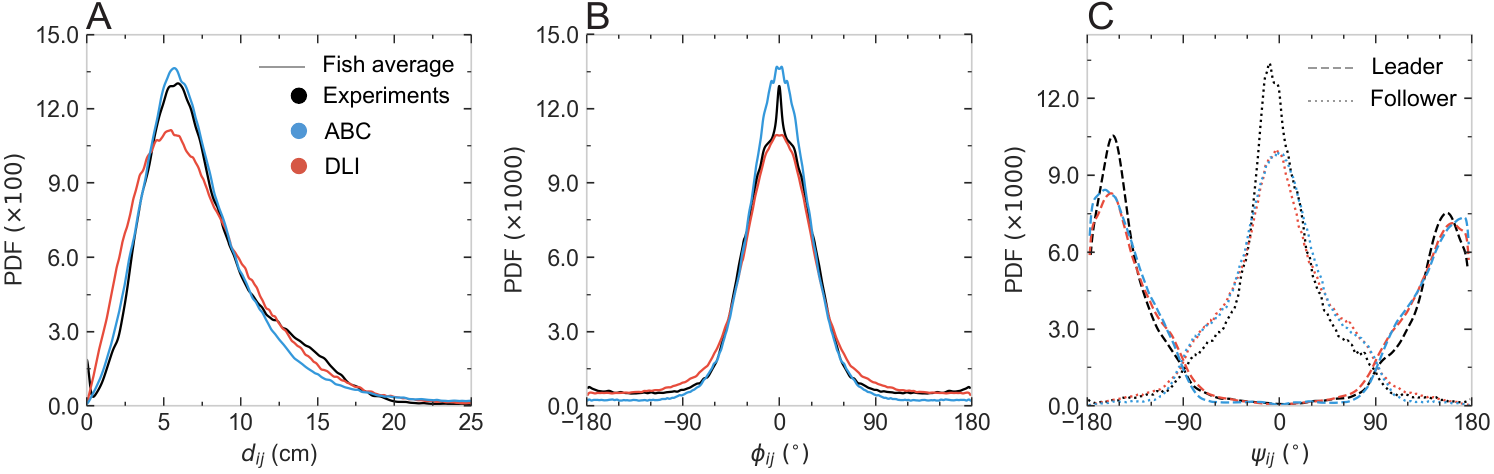}
    \caption{Probability density functions (PDF) of observables characterising collective behaviour: {\bf A} Distance between individuals $d_{ij}$, {\bf B} difference in heading angles $\phi_{ij}$, and {\bf C} angle of perception of the geometrical leader and follower $\psi_{ij}$. Black lines: experimental fish data. Blue lines: agents of the Analytical Burst-and-Coast model (ABC). Red lines: agents of the Deep Learning Interaction model (DLI). Only in panel {\bf C}, dashed lines: geometrical leader; dotted lines: geometrical follower.}
    \label{fig:collective_quantities_virtual}
\end{figure*}

\subsection{Quantification of the instantaneous individual behaviour}
\label{sec:individual}

The individual fish behaviour is characterised by three observables: the probability distribution function (PDF) of the speed $V$, of the distance to the wall $r_{\rm w}$, and of the angle of incidence to the wall $\theta_{\rm w}$. When swimming in pairs, fish tend to adopt a typical speed of about 7\,cm/s (see the peak of the PDF in Fig.~\ref{fig:individual_quantities_virtual}A), but can also produce high speeds up to 25--30\,m/s. In fact, we observe that both the leader and follower fish produce very similar speed profiles (thus omitted in Fig.~\ref{fig:individual_quantities_virtual}A). Both fish remain close to the wall of the tank (a consequence of the fish burst-and-coast swimming mode~\cite{calovi2018disentangling}), the leader being closer to the wall (typically, at about 0.5\,BL) than the follower (at about 1.2\,BL; see Fig.~\ref{fig:individual_quantities_virtual}B). This feature is due to the follower fish trying to catch up with the leader fish by taking a shortcut while taking the turn. Moreover, fish spend most of the time almost parallel to the wall: see the peaks of both PDFs at $\theta_{\rm w} \approx \pm 90^\circ$ in Fig.~\ref{fig:individual_quantities_virtual}C. A slight asymmetry is observed in the PDF of $\theta_{\rm w}$, showing that, in the experiments, fish have turned more frequently in the counter-clockwise direction. Values of the mean and the standard deviation of the PDFs presented in this section are given in Tables~S1, S2, and S3 in \nameref{SIText1}.

Both ABC and DLI models produce agents that move at the same mean speed as fish in the experiments, and Fig.~\ref{fig:individual_quantities_virtual}A shows that the speed PDF for both models are in excellent agreement with the one observed in real fish. Moreover, the agents of the ABC model are as close to the wall and as parallel to it as fish are. The PDF of the ABC leader is in good agreement with that of the fish leader
(Fig.~\ref{fig:individual_quantities_virtual}B). However, the PDF for the ABC follower has a peak at about the same distance to the wall as that of the leader, while the corresponding peaks are more separated for real fish. Yet, the PDF for the ABC follower is broader than for the leader, showing that the ABC follower tends to be farther from the wall than the leader, as observed for real fish. For the DLI model, the peaks of both leader and follower PDFs are at about the same position as for real fish, although their height is smaller than for fish, meaning that DLI-agents tend to explore more frequently the interior of the tank (observe the thicker tails of the PDF of $r_{\rm w}$ for the DLI model in Fig.~\ref{fig:individual_quantities_virtual}B). Alignment with the wall is also well reproduced by both models (Fig.~\ref{fig:individual_quantities_virtual}C), including the asymmetry in the direction of rotation around the tank: their peak at $\theta_{\rm w} > 0$ is higher than the one at $\theta_{\rm w} < 0$. As already seen in the PDF of $r_{\rm w}$, DLI-agents visit more often the interior of the tank, and are hence less aligned with the wall than the real fish and ABC agents. Note that the tendency of DLI-agents to rotate more frequently in the counterclockwise direction is learned from the training set, while this asymmetry has to be explicitly implemented in the ABC model, by introducing an asymmetric term in the analytical expression of the wall repulsion function.
A closer look at Fig.~\ref{fig:individual_quantities_virtual}C shows that fish actually follow the wall with a most likely angle of incidence
$\lvert\theta_{\rm w}\rvert$ that is slightly smaller \mbox{than 90$^\circ$}, a feature resulting from the burst-and-coast swimming mode inside a tank with positive curvature: fish are found more often going toward the wall than escaping it.

We have also computed the Hellinger distance (HD) between the experimental PDF probing the individual behaviour and the corresponding PDF produced by the DLI and ABC models. The Hellinger distance (see the caption of Tables~S10-S11, for more details) quantifies the (dis)agreement between two PDF for the same variable. The results of Tables~S10-S11 for both models confirm their good  performance: the DLI model HD is slightly better than that of the ABC model for the speed PDF, as good for the PDF of $r_{\rm w}$, and not quite as good for the PDF of $\theta_{\rm w}$.

\begin{figure*}[!ht]
    \centering
    \includegraphics[width=1.\linewidth]{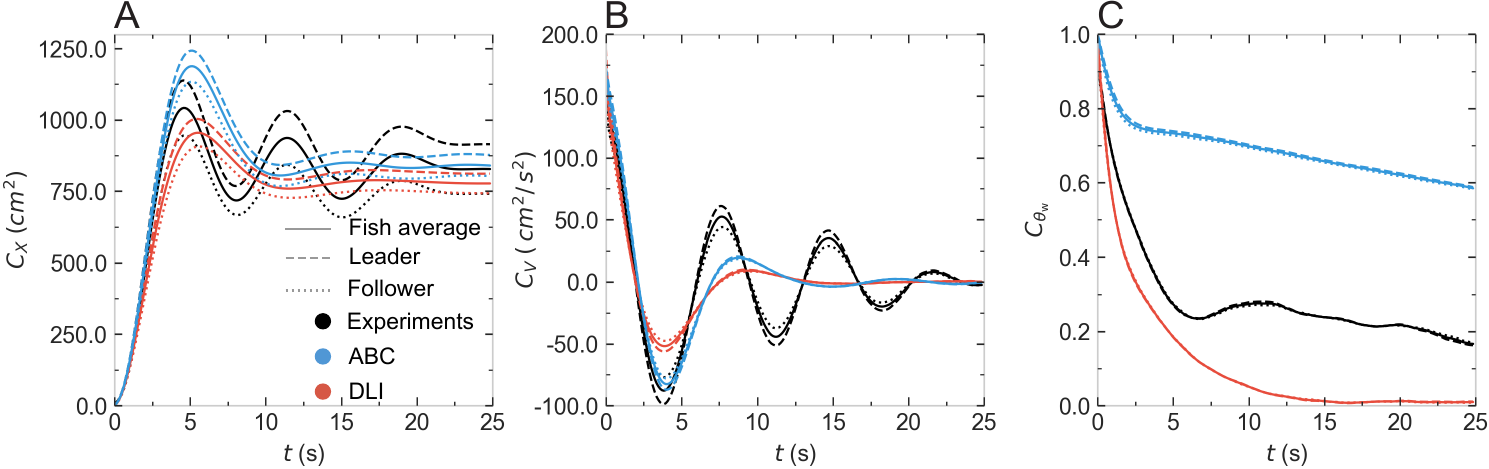}
    \caption{Observables quantifying temporal correlations in the system. {\bf A} Mean squared displacement $C_X(t)$, {\bf B} Velocity temporal autocorrelation $C_V(t)$, {\bf C} Temporal correlations of the angle of incidence to the wall $C_{\theta_{\rm w}}(t)$. Black lines: experimental fish data. Blue lines: agents of the Analytical Burst-and-Coast model (ABC). Red lines: agents of the Deep Learning Interaction model (DLI). Dashed lines: geometrical leader; dotted lines: geometrical follower; full lines: average over the 2 fish or agents.}
    \label{fig:correlation_quantities_virtual}
\end{figure*}

\subsection{Quantification of the instantaneous collective behaviour}
\label{sec:collective}

\textit{H.~rhodostomus} is a social species, and Fig.~\ref{fig:collective_quantities_virtual}A shows that the two fish remain most of the time close to each other, with the PDF of their distance $d_{ij}$ presenting a peak around $d_{ij}\approx 7\,{\rm cm}\approx 2$\,BL (mean and standard deviation of the PDFs presented in this section are given in Tables~S1, S2, and S3 in \nameref{SIText1}. The PDF of $d_{ij}$ produced by the DLI model is slightly wider than for the experiment and the ABC model, and in particular, presents too much weight at small distances.

The fish have a strong tendency to align with each other, as shown in Fig.~\ref{fig:collective_quantities_virtual}B, with the PDF of their relative heading $\phi_{ij}$ being sharply peaked at $0^\circ$. In addition, the PDF of the viewing angle $\psi_{ij}$ reveals that the fish are swimming one behind the other rather than side-by-side. This is illustrated in Fig.~\ref{fig:collective_quantities_virtual}C by the sharp difference in the PDF of the viewing angle for the leader and the follower.
The PDF of $\psi_{\rm leader} $ is peaked around $ \pm 160^\circ$, meaning that the follower fish is almost right behind the leader fish, but slightly shifted to the right or left. A slight left-right asymmetry in the PDF of the viewing angles is also visible, the follower being more frequently on the left side of the leader, a consequence of the fact that the fish in the experiment follow the wall by turning more often counterclockwise (Fig.~\ref{fig:individual_quantities_virtual}C), with the follower swimming farther from the wall than the leader (Fig.~\ref{fig:individual_quantities_virtual}B).

All these features are well reproduced by both models, with only some small quantitative deviations. The ABC model reproduces almost perfectly the experimental PDF of the distance between fish, whereas the PDF for the DLI model is only slightly wider and presents slightly more weight at very small distance than found for real fish or in the ABC model (Fig.~\ref{fig:collective_quantities_virtual}A). The DLI model is in turn better than the ABC model at reproducing the PDF quantifying the alignment of the fish, the latter producing more weight near $0^\circ$ than for real fish (Fig.~\ref{fig:collective_quantities_virtual}B). Both models fail at reproducing the small weight in the PDF at $\phi_{ij} \approx \pm 180^\circ$, which corres\-ponds to sudden U-turns that real fish sometime perform. The PDF of the viewing angles for the leader and the follower (Fig.~\ref{fig:collective_quantities_virtual}C) are also fairly reproduced by both models, including the slight left-right asymmetry observed in real fish, although the peak in the PDF at
$\psi_{\rm follower} = 0^\circ$ (and to a lesser extent at $\psi_{\rm leader} \approx - 160^\circ$) is not quite as sharp as in the experiment.

Again, we have computed the Hellinger distance between the experimental PDF probing the collective behaviour and the corresponding PDF produced by the DLI and ABC models. The results of Tables~S10-S11 for both models confirm their good  performance: as anticipated above, the DLI model HD for the PDF of the distance between agents is higher than for the ABC model (and is the highest found for all 6 PDF presented here, with ${\rm HD}_{d_{ij}} = 0.13$). However, Tables~S10-S11 also confirm that the DLI model reproduces quantitatively the PDF of $\phi_{ij}$ and $\psi_{ij}$.

\subsection{Quantification of temporal correlations}
\label{sec:correlations}

Fig.~\ref{fig:correlation_quantities_virtual} shows the 3 observables defined in Eqs.~(\ref{eq:cx}-\ref{eq:ctheta}) and probing the emerging temporal correlations in the system: the mean squared displacement $C_X(t)$, the velocity autocorrelation $C_V(t)$, and the autocorrelation of the angle of incidence to the wall $C_{\theta_{\rm w}}(t)$, as function of the time difference $t$ between observations. The figure reveals that both models fail to fully reproduce quantitatively these very non-trivial observables, which indeed constitute the most challenging benchmark characterising the correlations emerging from the fish behaviour.

Fish data present 3 distinct regimes: a quasi-ballistic regime at short timescale ($t\lesssim 1.5\,$s) where $C_X(t)\approx \langle v^2\rangle t^2$, followed by a second short diffusive regime ($1.5\,{\rm s}\lesssim t\lesssim 5$\,s) where $C_X(t)\approx D t$, which is limited by the finite size of the tank, ultimately leading to a third regime of saturation ($t > 5$\,s) cha\-rac\-te\-rised by slowly damped oscillations since fish are guided by the wall (Fig.~\ref{fig:correlation_quantities_virtual}A). Accordingly, the velocity correlation function starts from $C_V(t=0) = \langle v^2\rangle $ at short time and also presents damped oscillations (Fig.~\ref{fig:correlation_quantities_virtual}B). The negative minima of the oscillations in $C_V(t)$ correspond to times when the focal fish is essentially at a position diametrically opposite to its position at the reference time $t=0$, its velocity then being almost opposite to that at $t=0$. Similarly, positive maxima correspond to times when the fish returns to almost the same position it had at $t=0$, with a similar velocity, guided by the tank wall. Of course, these oscillations are damped as correlations are progressively lost, and the velocity correlation function $C_V(t)$ ultimately vanishes at large time $t\gg 20\,$s, due to the actual stochastic nature of the trajectories at this timescale (possible U-turns, or the fish randomly crossing the tank). Note that $C_X(t)$ is markedly different for the leader and follower fish, with a higher saturation value for the leader, which swims closer to the wall, as mentioned above.

The ABC model is able to fairly reproduce the short and intermediate regimes for $C_X(t)$ (Fig.~\ref{fig:correlation_quantities_virtual}A), as well as the position of its first peak, reached only 1\,s later than for fish. The ABC model also reproduces the experimental saturation value of $C_X(t)$ averaged over the two fish. As for the DLI model, its predictions are only slightly worse than that of the ABC model, since the DLI agents are moving a bit farther to the wall compared to ABC agents and real fish. Yet, both models equally fail at producing more than one oscillation, and the correlations are damped faster compared to the experiment.

As for the velocity autocorrelation $C_V(t)$ (Fig.~\ref{fig:correlation_quantities_virtual}B), the ABC model reproduces almost perfectly the short and intermediate regimes and the position of the first negative minimum (hence, up to $t=6$\,s), while the DLI model underestimates the depth of this first minimum. But again, both models fail at reproducing the persistence of the correlations, producing a too fast damping of the oscillations (an effect slightly stronger in the DLI model).

Both models struggle at reproducing the correlation function $C_{\theta_{\rm w}}(t)$ of the angle of incidence to the wall (Fig.~\ref{fig:correlation_quantities_virtual}C), where the fish curve first sharply decreases up to $t=6$\,s and then remains close to $C_{\theta_{\rm w}}\approx 0.2$. The ABC model is clearly unable to reproduce both the decreasing range (clearly diverging before $t=2$\,s) and the correct saturation value (never falling below $C_{\theta_{\rm w}}\approx 0.6$). As for the DLI model, it produces a slightly sharper decay of $C_{\theta_{\rm w}}(t)$ than for real fish, up to $t \approx 6$\,s, but fails to reproduce the non-negligible remaining persistence of the correlation observed in fish for $t > 7\,$s, with $C_{\theta_{\rm w}}(t)$ in the DLI model decaying rapidly to zero. In fact, both models fail to reproduce the experimental $C_{\theta_{\rm w}}(t)$ for opposite reasons. The ABC model exhibits a too high persistence of the correlations of $\theta_{\rm w}$ compared to real fish, presumably because real fish indeed often follow the wall but can also produce sharp U-turns, as observed in Fig.~\ref{fig:individual_quantities_virtual}C. On the other hand, the failure of the DLI model in reproducing $C_{\theta_{\rm w}}(t)$ stems from the fact that DLI agents move farther from the wall and cross through the tank more often than real fish and ABC agents (see the discussion of Fig.~\ref{fig:individual_quantities_virtual}B above), hence leading to a too fast, and ultimately total, loss of correlation for $\theta_{\rm w}$.

\subsection{Complementary analyses}
\label{sec:complementary}

In order to test whether the DLI model has correctly learned the presence of the wall, we have run 30 simulations of duration 6000\,s to check whether the DLI agents would stay within the area of the tank, even without enforcing its presence by the rejection procedure mentioned in the second paragraph below Eq.~(\ref{eq:motion_eqs2}). We found that the DLI agents indeed remain in or very near the tank during the entire time of the simulation in 60\,\% of runs. In the other 40\,\% of runs, the DLI agents would ultimately escape the tank after a mean time of order 3000\,s. These results are summarised in Fig.~S1 of \nameref{SIText1}, where we present the time series of the distance to the wall $r_{\rm w}(t)$ for the 10 first runs, and in  Fig.~S2 of \nameref{SIText1}, where we report the survival probability (\textit{i.e.}, the probability that the DLI agents remain within the tank up to a given time). These results indicate that the DLI model has convincingly learned the presence of the wall, and is able to maintain the agents within the wall for several dozen of minutes without the need of an explicit rejection procedure.

We have also conducted several other complementary tests of our approach. First, the DLI model yields better results in generating social interactions than a similarly purposed ANN for human trajectory forecasting~\cite{alahi2016social, kothari2020human}  (D-LSTM model; see Fig.~S3 and S4 in \nameref{SIText1}, Tables~S4, S5, S6, S12  in \nameref{SIText1}, and \nameref{mov:dlstm}). In particular, the results for the Hellinger distance (${\rm HD}_{{r_{\rm w}}} = 0.30$ and ${\rm HD}_{{\theta_{\rm w}}} = 0.40$) show that this D-LSTM model completely fails at capturing the interaction of the fish with the tank wall.
While this is expected due to the missing inputs (compared to the DLI; see \nameref{SIText1}), these results confirm that there exist models that do indeed capture the short-term dynamics without being able to reproduce the long-term dynamics, presumably due to non-Markovian effects. In addition, we also trained a Multi-layered Perceptron Interaction (MLI) model \textit{without any memory cells}, and found that it fails to reproduce all 6 PDF (see \nameref{SIText1}, Fig.~S6), resulting in high values of the corresponding Hellinger distances (see \nameref{SIText1}, Table~S13).

Moreover, we have analysed the performance of the DLI model when varying the fraction of the dataset used in its training. The performance is quantified by using the Hellinger distance (HD) between the experimental PDF and that produced by the DLI model, and \nameref{SIText1}, Table~S15 reports the resulting  HD values.
When only using 75\%, 50\%, or even 37.5\% of the dataset, the DLI model has a similar performance as when trained with the full dataset (4 hours of pair trajectories). However, the performance sharply drops when only using 25\%, 12.5\%, and 5\% of the dataset. In fact, using 25\% or less of the dataset, we also found that the performance significantly depends on the training sample (we ran 4 training sessions in each case). Finally, we also found that without enforcing the presence of the wall with our rejection procedure, the median escape time of the fish computed over 30 runs of 6000\,s  when using 25\%, 12.5\%, and 5\% of the dataset are of order 500\,s, 75\,s, 50\,s, compared to 3000\,s when using 100\% or even 50\% of the dataset.
These results show that our DLI network (and its size) is coherent with the size of the training dataset, and that its predictions remain robust when restricting the data at least down to half of the original dataset.

Finally, we have trained the DLI model with data for pairs of zebrafish (\textit{D.~rerio}), and found that it yields fair results for this species too, without any structural modification in its architecture (see Fig.~S5 and Tables~S8, S9, and S14 in \nameref{SIText1}). While acquiring a functional model of a new species' interactions proved straightforward with the DLI, the same would not be generally true for analytical models.

Following the completion of the present work, we have exploited the DLI model to study groups with more than two fish, \textit{without any retraining}. Indeed, \textit{H.~rhodostomus}~\cite{lei2020computational}, like many other group-living species~\cite{cavagna2018physics}, effectively only interact with a few influential neighbours, at a given time. Thus, for a given agent in a group of $N>2$ agents, the DLI for \textit{H.~rhodostomus} should only retain the influence of typically the two agents leading to the highest acceleration~\cite{wang2022impact, lei2020computational}, as predicted by the DLI model. \nameref{mov:dli5} illustrates this procedure for $N=5$ agents, resulting in a cohesive and aligned group, in qualitative agreement with experimental observation~\cite{lei2020computational}. In addition, the present DLI model has also been recently exploited in \cite{papaspyros2023quantifying} to command a robot fish initially introduced in~\cite{papaspyros2023biohybrid} (where it was commanded by the ABC model), and moving alone in the tank, or reacting in a closed-loop to 1 or 4 real fish.

\section{Conclusions \& Discussion}

Studying social interactions in animal groups is crucial to understand how complex collective behaviours emerge from individuals' decision-making processes. Very recently, such interactions have been extensively investigated in the context of collective motion by exploiting classical computational modelling~\cite{calovi2018disentangling, escobedo2020data, jayles2020collective} and automated machine learning-based methods~\cite{heras2018deep, costa2020automated}. Although ML algorithms have been shown to provide insight into the interactions of hundreds of individuals at short timescales~\cite{heras2018deep, costa2020automated}, their ability to reproduce the complex dynamics in animal groups at long timescales has not yet been assessed.

Here, we have presented a deep learning interaction model (DLI) which reproduces the behaviour of fish swimming in pairs. The DLI model good performance can be primarily ascribed to its memory related to a biologically relevant timescale (fish kicks of typical duration 0.5--0.6\,s), and to a carefully crafted input/feature vector. Indeed, the MLI model without memory cells performs very poorly, while the D-LSTM model, characterised by a different input/feature vector, demonstrates markedly lower performance than the DLI model. 

We have also introduced the appropriate tools for the validation of an ANN model, when compared to experimental results and confronted with an analytical behavioural model (ABC). In fact, our study establishes a systematic methodology to assess the long-term predictive power of a model (analytical or ML), by introducing a set of fine observables probing the individual and collective behaviour of model agents, as well as the subtle correlations emerging in the system. These observables, which can be straightforwardly extended to groups of $N>2$ agents, provide an extremely stringent test for any model aimed at producing realistic long-term trajectories mimicking that of actual animal groups. In particular, we consider that the usual validation of an ML model at a short timescale should be complemented by the type of long timescale analysis that we propose here, in order to fully assess its performance. Indeed, we have shown that a model (like the D-LSTM model) can have a good performance at very short timescales, while presenting a degraded performance at large timescales, presumably due to non-trivial non-Markovian effects.

The DLI model closely reproduces the dynamics of real fish at both the individual (speed, distance to the wall, angle of incidence to the wall) and collective (distance between in\-di\-vi\-duals, relative heading angle, angle of perception) levels during long simulations corresponding to more than 16~hours of fish swimming in a tank, hence successfully generating life-like interactions between agents. When compared to experiment, the ABC model and the DLI model essentially performs equally well. Notably, the DLI model better captures the most likely distance of the leader and follower from the wall. However, the DLI model is less accurate in reproducing the temporal correlations quantified by the mean-squared displacement and the velocity autocorrelation. Yet, both ABC and DLI models fail at capturing the temporal correlations of the angle of incidence to the wall, but for very different reasons. More importantly, the DLI model convincingly infers the presence of the tank wall, and is able to keep the DLI agents within the wall boundaries for several dozen minutes, even when the rejection procedure is not enforced. In addition, we have shown that the performance of the DLI model remains robust even when only using half of the experimental training dataset, while its accuracy sharply drops when only using a quarter of the training dataset.

Our study demonstrates two advantages of ML techniques: 1)~they can drastically accelerate the generation of new models (as illustrated here for zebrafish), and 2)~with minimal expertise in biology or modelling. This is especially useful in robotics, where models often act as behavioural controllers (\textit{i.e.}, trajectory generators) that guide the robot(s). Although there already exist many bio-hybrid experiments in the literature, most of them rely on simplified models for behavioural modulation~\cite{papaspyros2019bidirectional, bonnet2016infiltrating, bonnet2018closed}, few of them exploit realistic models (analytical or ML)~\cite{cazenille2018mimetic, cazenille2018blend}, and, to our knowledge, none of them are tested in the long term in simulations or real-life. In this context, ML has the potential to benefit multidisciplinary studies, provided such techniques are thoroughly validated in simulations.

However, accelerating the production of collective behaviour models with ML comes at a cost. Indeed, the DLI is a black-box model, and although it captures the subtle impact of social interactions between individuals, it is impossible to retrieve the interaction functions themselves. Some approaches partially address this issue by providing insight into how the network operates for specific sets of inputs~\cite{heras2018deep, costa2020automated}. Yet, they still do not offer explicit interaction functions. 
Instead, they provide insights in the form of force maps that can, to some extent, be used to interpret the underlying mechanisms of the interactions, or in the form of input/output correlation graphs, that showcase the manner in which an input state typically affects the output \cite{lundberg2017unified}.
On the other hand, analytical models supplemented by a procedure to reconstruct social interactions~\cite{calovi2018disentangling, escobedo2020data} provide a concise and explicit description of the system in question. Moreover, varying the parameters of such models allows for investigating their relative impact on the dynamics, in the form of phase diagrams representing the collective observables (and the corresponding collective state of the group) as a function of the model parameters~\cite{calovi2014swarming, wang2022impact}. This is not feasible with ML models, unless they are retrained or specifically structured to allow it.

In summary, this work shows that DLI-like models may now be considered as firm candidates to shed light on groundbreaking problems such as how social interactions take place and affect collective behaviour in living groups. Yet, we have emphasised that social interaction models should be precisely tested at both short \textit{and} long timescales. Future work includes the design of ANNs that provide additional information about the learned dynamics (\textit{e.g.}, using the framework of~\cite{lundberg2017unified} and/or attention layers, like in~\cite{heras2018deep, costa2020automated}), or possibly, by exploiting symbolic regression algorithms~\cite{quade2016prediction, chen2019revealing}. We also plan to study the extension of the DLI model to larger groups, in particular, in connection with our robotic platform~\cite{papaspyros2019bidirectional, bonnet2016infiltrating, bonnet2018closed,papaspyros2023biohybrid,papaspyros2023quantifying}. It would also be interesting to apply the DLI model in different environmental conditions, such as light intensity, as recently done for the ABC model~\cite{xue2023tuning}. Ultimately, a more generalised and unified version of the DLI model or similar algorithms requires extensive testing with additional social animal species (\textit{e.g.}, humans). We believe that these approaches could improve our understanding of the mechanisms arising in collective behaviour and allow for more precisely exploring and modulating them.


\enlargethispage{20pt}

\ethics{The experiments conducted with \textit{H.~rhodostomus} were approved by the Ethics Committee for Animal Experimentation of the Toulouse Research Federation in Biology no. 1 and comply with the European legislation for animal welfare. The experiments conducted with \textit{D.~rerio} were approved by the state ethical board of the Department of Consumer and Veterinary Affairs of the Canton de Vaud (SCAV) of Switzerland (authorisation no. 2778).}

\dataccess{All the code concerning the data pre-processing, neural networks, and plot scripts is publicly available at \url{https://doi.org/10.5281/zenodo.7634912}. Experimental and generated data are available at \url{https://doi.org/10.5281/zenodo.7634687}.}

\aucontribute{F.M.,~V.P.,~C.S.,~G.T conceived and designed the study. V.P. and R.E implemented the data pre-processing. V.P. implemented the neural networks, plot scripts, and concept figures. C.S.~also assisted with the plot scripts and implemented the simulation of the model from~\cite{calovi2018disentangling}. The design and fine-tuning of the neural network structures was done by V.P. Moreover, A.A.~and V.P. configured the D-LSTM. V.P.~and~C.S.~analysed the data. C.S., G.T. conducted the experiments and their analysis with \textit{H.~rhodostomus}, and V.P.~conducted the experiments with \textit{D.~rerio}. V.P.~and~C.S wrote the original paper draft, and A.A.,~R.E.,~F.M.,~V.P.,~C.S.,~G.T. reviewed and edited the draft, and produced the final version.}

\competing{The authors declare no competing interests.}

\funding{This work was partly supported by the Germaine de Staël project no. 2019-17. V.P. was also supported by the Swiss National Science Foundation project `Self-Adaptive Mixed Societies of Animals and Robots', grant no. 175731. R.E., G.T.~and C.S. were supported by the French National Research Agency (ANR-20-CE45-0006-01). G.T. also acknowledges the support of the Indo–French Centre for the Promotion of Advanced Research (project N°64T4-B) and gratefully acknowledges the Indian Institute of Science to serve as Infosys visiting professor at the Centre for Ecological Sciences in Bengaluru.}

\ack{We would like to thank Dr.~Frank Bonnet for his contribution in the early stages of this paper and the Franco-Swiss collaboration between the \'Ecole Polytechnique F\'ed\'erale de Lausanne (V.P.~and F.M.), the Centre de Recherches sur la Cognition Animale (R.E.~and G.T.), and the Laboratoire de Physique Th\'eorique (C.S.), the last two at Universit\'e de Toulouse III -- Paul Sabatier.}
\end{multicols}



\printbibliography

\section*{Supporting information}


\section*{SI Text}

\paragraph*{Text~S1} 
\label{SIText1}
This PDF file presents simulations of the DLI model without enforcing the presence of the tank wall, a comparison of the DLI to a similarly purposed neural network, and simulation results with neural networks that do not have memory cells. It also presents the validation of the DLI’s scalability to another fish species. Furthermore, it  addresses the performance of the DLI model with respect to the fraction of the data considered in its training. Finally, this file reports tables for the Hellinger distance associated with these different conditions. This document includes 6 figures and 15 tables.

\section*{SI Videos}

\paragraph*{Video~S1.}
\label{mov:comparison}
Examples of trajectories obtained in  experiments with \textit{H. rhodostomus} (left), for the Analytical burst-and-coast (ABC) model (centre), and for the Deep Learning Interaction (DLI) model (right). This video illustrates the qualitative agreement between trajectories generated by the ABC and DLI models and experimental trajectories, while the quantitative agreement between the models and experiments is studied in detail in the Result section. The video can be downloaded at \url{https://github.com/epfl-mobots/preddl_2023/tree/v1.0.5/Videos/S1_Video.mp4}.

\paragraph*{Video~S2.}
\label{mov:dlstm}
Example of a generated trajectory simulation for the D-LSTM model. Already at the qualitative level, the D-LSTM model fails at reproducing realistic trajectories (compare with \nameref{mov:comparison}). The video can be downloaded at \url{https://github.com/epfl-mobots/preddl_2023/tree/v1.0.5/Videos/S2_Video.mp4}.

\paragraph*{Video~S3.}
\label{mov:dli5}
Example of collective behaviour in a group of 5 DLI agents, \textit{without any retraining}. For a given focal agent, we compute the predicted acceleration and noise which would be produced by each of the 4 other agents. Following~\cite{lei2020computational}, we define the two most influential neighbours as the neighbours leading to the two highest predicted accelerations. Ultimately, the focal fish speed and position are updated according to Eqs.~(9-11), using the sum of these two highest accelerations and the average predicted noise. This video illustrates the fact that, although the DLI was only trained to mimic the social interactions between pairs of fish, it produces cohesive and aligned trajectories for 5 agents, in good qualitative agreement with corresponding trajectories for 5 rummy-nose tetra ~\cite{lei2020computational}. In the future, we will address the quantitative comparison between long-term trajectories for groups of DLI agents and real fish, in particular, in connection to our robotic platform~\cite{papaspyros2023biohybrid}. The video can be downloaded at \url{https://github.com/epfl-mobots/preddl_2023/tree/v1.0.5/Videos/S3_Video.mp4}.

\paragraph*{Video~S4.}
\label{mov:mli}
Example of a generated trajectory simulation for the Multi-layered Perceptron Interaction (MLI) model. Already at the qualitative level, the MLI model fails at reproducing realistic trajectories (compare with \nameref{mov:comparison}). The video can be downloaded at \url{https://github.com/epfl-mobots/preddl_2023/tree/v1.0.5/Videos/S4_Video.mp4}.

\end{document}